\documentclass[11pt]{article}

\usepackage[final]{acl}

\usepackage{times}
\usepackage{latexsym}

\usepackage[T1]{fontenc}

\usepackage[utf8]{inputenc}

\usepackage{microtype}

\usepackage{inconsolata}

\usepackage{amsmath}
\usepackage{amssymb}

\usepackage{array}
\usepackage{booktabs}
\usepackage{multirow}
\usepackage[table]{xcolor}
\usepackage{arydshln} 

\usepackage{algorithm}
\usepackage{algorithmic}

\usepackage{tcolorbox}
\tcbuselibrary{breakable,skins,listings}
\usepackage{listings}
\usepackage{graphicx}
\usepackage{subcaption}

\usepackage{CJKutf8}
\renewcommand{\foreignlanguage}[2]{\begin{CJK}{UTF8}{gbsn}#2\end{CJK}}

\title{Better Literary Translation: A Multi-Aspect Data Generation and \\LLM Training Approach}

\author{%
Zhihao Lin$^1$ \quad Ziqi Zhu$^1$ \quad Hao Huang$^1$ \quad Guanghui Wang$^1$ \quad Peiyang He$^{*1,2}$\\
  $^{1}$Amazon Web Services (AWS) \quad $^{2}$Peking University\\
\texttt{\{lzhihao, ziqizhu, tonyhh, guanghu, peiyan\}@amazon.com}\\
}

\begin{document}

\maketitle

\renewcommand{\thefootnote}{\fnsymbol{footnote}}
\footnotetext[1]{Corresponding author.}
\renewcommand{\thefootnote}{\arabic{footnote}}

\begin{abstract}
Literary translation poses unique challenges due to the scarcity of high-quality annotated data and the need to balance expression fluency with literary effect.
We present a multi-aspect iterative refinement framework that generates high-quality translation references and preference data through specialized LLM translators, each targeting a distinct quality dimension.
We leverage the generated data for supervised fine-tuning and reinforcement learning.
Experiments show that our generated references outperform the original ground truth for SFT by 8.65 CEA100 points.
For reinforcement learning, we find that DPO leads to performance degradation in this setting, while leveraging an explicit reward model for GRPO yields an additional 1.51 point improvement.
We attribute this to the stability of two-stage training and GRPO's online exploration capability.
Our resulting models, LitMT-8B and LitMT-14B, achieve 67.25 and 69.07 CEA100 respectively on the MetaphorTrans English-to-Chinese literary translation benchmark, competitive with Claude Sonnet 4.5 at 68.43, and demonstrate strong generalization to out-of-domain literary work (\emph{i.e.}, O.\ Henry).
\end{abstract}

\section{Introduction}

Literary translation requires balancing expression fluency with literary effect.
Unlike general-domain translation, literary texts demand faithful rendering of metaphors, cultural nuances, and stylistic devices~\citep{matusov2019challenges,guerberof2022creativity}, creating inherent trade-offs between naturalness and artistic preservation.
High-quality literary translation enables readers across languages to experience the full artistic intent of original works---a capability of growing industrial importance, yet professional human translation remains costly and slow to scale.
While large language models have shown promising results on general translation tasks~\citep{kocmi2024findings,kocmi-etal-2025-findings}, literary translation remains challenging due to both the scarcity of high-quality training data and the difficulty of capturing nuanced quality trade-offs.

\begin{figure*}[t]
\centering
\begin{subfigure}[b]{0.32\textwidth}
    \centering
    \includegraphics[width=\textwidth]{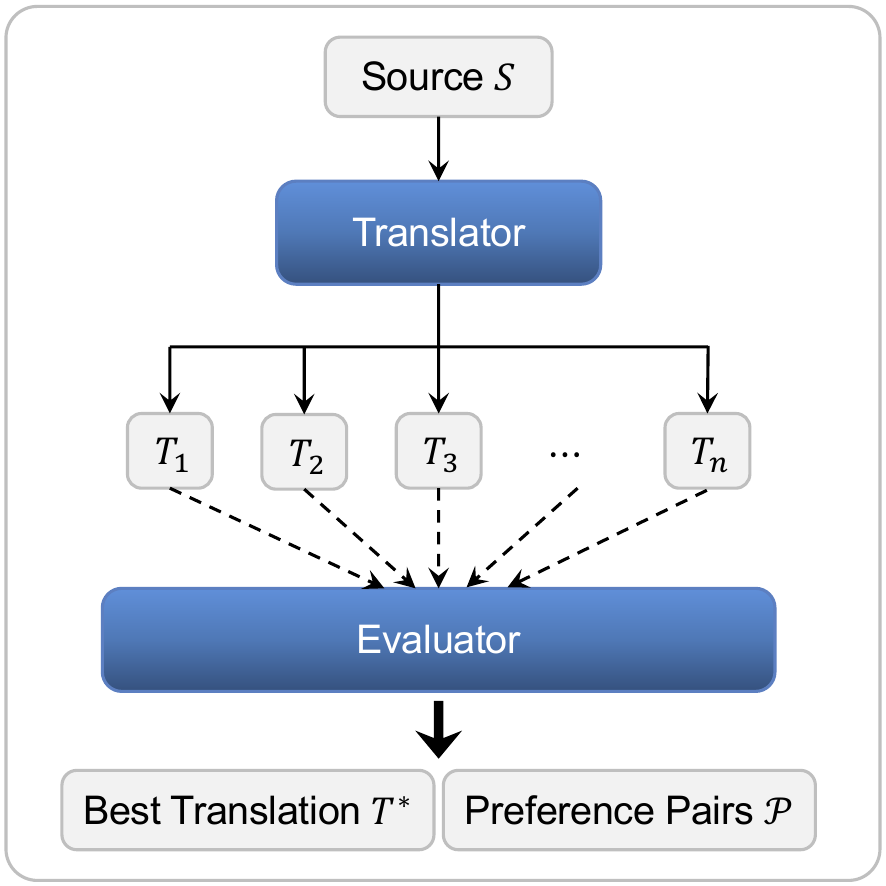}
    \caption{Sampling + LLM Judge}
    \label{fig:comparison-a}
\end{subfigure}
\hfill
\begin{subfigure}[b]{0.32\textwidth}
    \centering
    \includegraphics[width=\textwidth]{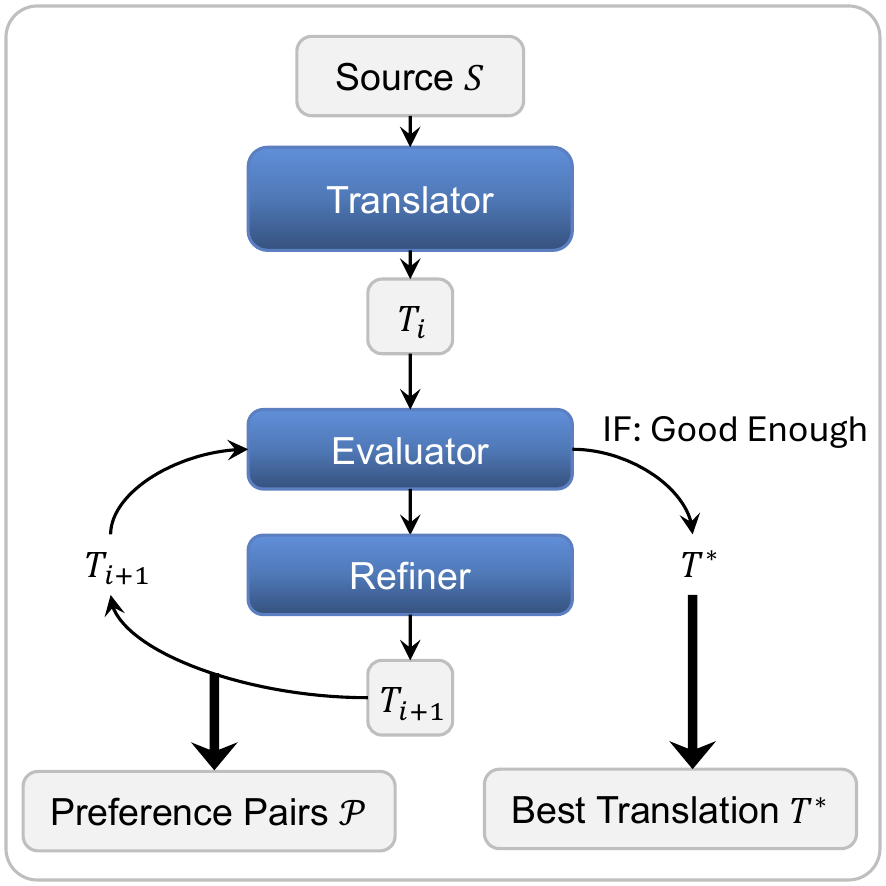}
    \caption{Self-Refinement}
    \label{fig:comparison-b}
\end{subfigure}
\hfill
\begin{subfigure}[b]{0.32\textwidth}
    \centering
    \includegraphics[width=\textwidth]{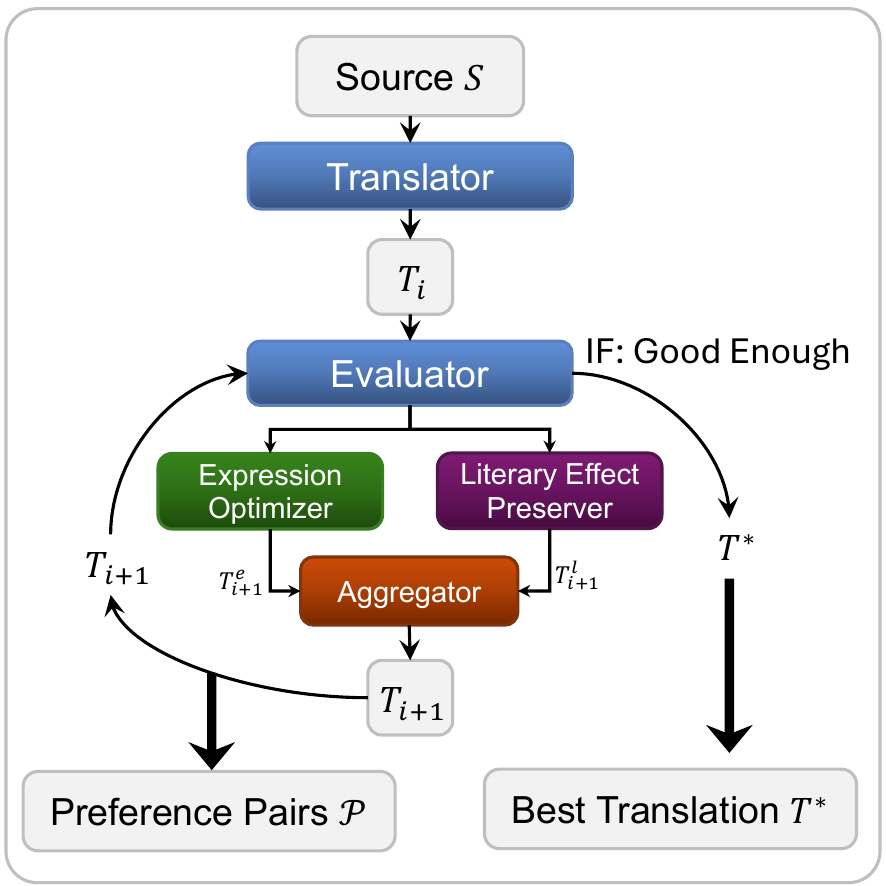}
    \caption{Ours: Multi-Aspect Refinement}
    \label{fig:comparison-c}
\end{subfigure}
\caption{Comparison of data generation approaches for literary translation. (a) \textbf{Sampling + LLM Judge}: samples multiple translations and selects the best via LLM scoring. (b) \textbf{Self-Refinement}: iteratively refines translations using a single LLM. (c) \textbf{Ours: Multi-Aspect Refinement}: decomposes quality into expression fluency and literary effect with specialized modules, generating both high-quality references and preference pairs.}
\vspace{-7pt}
\label{fig:comparison}
\end{figure*}

Recent approaches address this by training models with long chain-of-thought (CoT) reasoning~\citep{wang2025drt,wang2026deeptrans,wang2025extrans} or using massive LLMs as judges during reinforcement learning~\citep{zheng2023llmjudge,wang2026deeptrans,wang2025extrans}.
While effective, these methods face critical limitations:
(1) using LLM-as-a-judge as the reward signal requires expensive inference for every training sample, and this cost must be repeated for each new model, leading to prohibitive training costs;
(2) long CoT reasoning increases inference latency by up to 10$\times$, making these approaches impractical for real-time applications.

In this work, we propose a multi-aspect iterative refinement framework that generates both high-quality translation references and preference data.
Unlike approaches that require costly LLM inference for every training sample, our generated references and preference pairs can be reused across multiple training runs.
Our key insight is that literary translation quality can be decomposed into two dimensions---expression fluency and literary effect---which often conflict.
By employing specialized LLM translators that each focus on optimizing a single dimension, we generate high-quality references through iterative refinement, while the intermediate outputs naturally form preference pairs that capture quality trade-offs (Figure~\ref{fig:comparison}c).
This multi-aspect data provides effective targets for SFT. For preference optimization through RL training, we compare implicit reward modeling methods (\emph{e.g.}, DPO~\citep{rafailov2023direct}) with explicit reward modeling~\citep{ouyang2022training,eisenstein2023helping} followed by GRPO~\citep{shao2024deepseekmath}, and find that training a compact reward model on our data with GRPO significantly outperforms DPO-based approaches. Since our trained models do not rely on long CoT reasoning, they remain practical for real-time applications.

Our contributions are as follows:
\begin{itemize}
    \item We propose a \textbf{multi-aspect iterative refinement pipeline} for literary translation data generation. By decomposing translation quality into expression fluency and literary effect with specialized modules, our pipeline simultaneously produces high-quality translation references and preference pairs that capture the trade-offs between these dimensions.
    \item We conduct a systematic comparison between implicit reward modeling (\emph{e.g.}, DPO) and explicit reward modeling with GRPO for literary translation LLM training. We find that combining a learned reward model with BLEU and format constraints as a composite reward for GRPO significantly outperforms DPO-based methods.
    \item Our resulting models, \textbf{LitMT-8B} and \textbf{LitMT-14B}, achieve 67.25 and 69.07 CEA100 respectively on the MetaphorTrans English-to-Chinese literary translation benchmark, competitive with Claude Sonnet 4.5 (68.43), and demonstrate strong generalization to out-of-domain literary work (\emph{i.e.}, O.\ Henry).
\end{itemize}

\section{Related Work}

\subsection{Fine-Tuning LLMs for Literary Translation}

Recent work has fine-tuned specialized LLMs for literary translation to address both the high inference cost of commercial models and their inconsistent handling of literary devices such as metaphors and personification.
DRT~\citep{wang2025drt} synthesizes long chain-of-thought data through general LLMs for SFT.
DeepTrans~\citep{wang2026deeptrans} and ExTrans~\citep{wang2025extrans} apply reinforcement learning with LLM-as-a-judge~\citep{zheng2023llmjudge} (\emph{e.g.}, DeepSeek-V3) as the reward signal.
While these methods achieve strong results, they require expensive LLM inference for every training sample.
In contrast, our work focuses on generating high-quality training data---both references and preference pairs---that can be reused across various training paradigms without relying on massive LLMs during training.

\subsection{Data Generation for LLM Training}

Early approaches like Self-Instruct~\citep{wang2023self} and Alpaca~\citep{alpaca} use LLM outputs directly as SFT targets, producing single-point outputs without quality comparisons.
Rejection sampling methods such as Best-of-N~\citep{stiennon2020learning,bai2022training} and RSO~\citep{liu2023statistical} generate multiple outputs and select the best; however, one-shot sampling without iterative refinement often yields suboptimal results.
Multi-agent approaches like MATRIX~\citep{tang2025synthesizing} and Fellowship of the LLMs~\citep{arif2025fellowship} show promise for data generation, but have not been tailored for literary translation or addressed trade-offs between quality dimensions.
Our multi-aspect approach assigns each translator a specific dimension (expression fluency vs.\ literary effect), generating preference pairs that capture these trade-offs.

\section{Approach}

Our approach consists of two stages: a data generation pipeline that produces high-quality literary translations and preference pairs (\S\ref{sec:refinement}), followed by a training pipeline that leverages these outputs for model optimization (\S\ref{sec:training}).

\subsection{Multi-Aspect Iterative Refinement}
\label{sec:refinement}

\subsubsection{Pipeline Overview}

Figure~\ref{fig:comparison}~(c) illustrates our pipeline. Given a source text $S$, we first generate an initial translation $T_0$ using a naive translator. We then iteratively refine through the following loop: (1) an evaluator scores the current best translation and provides feedback identifying specific issues; (2) two specialized LLM translators---Expression Optimizer and Literary Effect Preserver---each generate an improved version based on the feedback; (3) an aggregator synthesizes both versions into a single translation; (4) the evaluator scores the aggregated result and updates the best translation if the score improves. The process terminates when the score exceeds a threshold, the iteration count reaches the maximum, or the score shows no improvement for consecutive rounds.

Algorithm~\ref{alg:refinement} formalizes this procedure.

\begin{algorithm}[t]
\small
\caption{Multi-Aspect Iterative Refinement}
\label{alg:refinement}
\begin{algorithmic}[1]
\REQUIRE Source $S$, threshold $\tau$, max rounds $K$, patience $N$
\ENSURE Best translation $T^*$, preference pairs $\mathcal{P}$
\STATE $T_0 \gets \textsc{Translate}(S)$ \COMMENT{Naive translator}
\STATE $s_0, F_0 \gets \textsc{Eval}(S, T_0)$ \COMMENT{Score and feedback}
\STATE $T^* \gets T_0$; $s^* \gets s_0$; $F^* \gets F_0$; $n \gets 0$
\FOR{$i = 1$ to $K$}
    \STATE $T_i^e \gets \textsc{ExprOpt}(S, T^*, F^*)$ \COMMENT{Fluency-focused}
    \STATE $T_i^l \gets \textsc{LitPres}(S, T^*, F^*)$ \COMMENT{Literary-focused}
    \STATE $T_i \gets \textsc{Aggregate}(S, T_i^e, T_i^l)$ \COMMENT{Synthesize}
    \STATE $s_i, F_i \gets \textsc{Eval}(S, T_i)$
    \IF{$s_i > s^*$}
        \STATE $T^* \gets T_i$; $s^* \gets s_i$; $F^* \gets F_i$; $n \gets 0$
    \ELSE
        \STATE $n \gets n + 1$
    \ENDIF
    \IF{$s^* \geq \tau$ \OR $n \geq N$}
        \STATE \textbf{break}
    \ENDIF
\ENDFOR
\STATE $\mathcal{P} \gets \{(T_w, T_l) : s_w > s_l\}$ \COMMENT{All evaluated translations}
\RETURN $T^*$, $\mathcal{P}$
\end{algorithmic}
\end{algorithm}

\subsubsection{Module Design}

We now describe each module in detail. Our pipeline consists of five modules, all implemented as LLM prompts (see Appendix~\ref{app:data_gen_prompts} for details).

\noindent\textbf{Naive Translator.} Generates an initial literary translation $T_0$ from source $S$ without emphasis on any particular quality dimension, providing a neutral starting point for refinement.

\noindent\textbf{Expression Optimizer.} Focuses on target-language fluency: natural word order, idiomatic collocations, and concise expression. Given source $S$, current best translation $T^*$, and feedback $F^*$, it generates a fluency-optimized version $T_i^e$.

\noindent\textbf{Literary Effect Preserver.} Focuses on preserving artistic qualities: figurative language, rhetorical devices, and tone. Given the same inputs $(S, T^*, F^*)$, it generates a literary-focused version $T_i^l$.

\noindent\textbf{Aggregator.} Synthesizes improvements from both specialized modules. Given source $S$ and the two refined translations $(T_i^e, T_i^l)$, it produces a final literary translation $T_i$ that balances expression fluency with literary effect.

\noindent\textbf{Evaluator.} Scores literary translations based on expression fluency and literary effect, and provides textual feedback identifying specific issues to guide the refinement modules.

\subsubsection{Outputs}

The pipeline produces two types of training data that directly feed into our training pipeline (\S\ref{sec:training}).

\noindent\textbf{High-Quality References.} For each source, we select the best translation $T^* = \arg\max_t \text{score}(t)$ across all iterations. These serve as targets for supervised fine-tuning.

\noindent\textbf{Preference Pairs.} From the iteration history, we construct preference pairs by enumerating all translation pairs $(T_w, T_l)$ where $T_w \neq T_l$ and $\text{score}(T_w) > \text{score}(T_l)$.
This yields multiple pairs per source, capturing fine-grained quality distinctions.
These pairs support both direct preference optimization (\emph{e.g.}, DPO~\citep{rafailov2023direct}, CPO~\citep{xu2024contrastive}, SimPO~\citep{meng2024simpo}) and reward model training~\citep{ouyang2022training}.

\subsection{Training Pipeline}
\label{sec:training}

We explore three training strategies using the generated data: supervised fine-tuning with high-quality references, and two RL approaches---implicit reward modeling (\emph{e.g.}, DPO, SimPO, CPO) and explicit reward modeling (RM+GRPO).

\subsubsection{Supervised Fine-Tuning}

As a baseline, we train on the best translations $y^*$ using cross-entropy loss:
\begin{equation}
\small
\mathcal{L}_{\text{SFT}} = -\mathbb{E}_{(x,y^*)} \left[\log \pi_\theta(y^*|x)\right].
\end{equation}

\subsubsection{Implicit Reward: DPO-Series}

DPO-series methods directly optimize policy using preference pairs without training a separate reward model.
Taking DPO as an example:
\begin{equation}
\small
\mathcal{L}_{\text{DPO}} = -\mathbb{E} \left[\log \sigma\left(\beta \log \frac{\pi_w}{\pi_w^{\text{ref}}} - \beta \log \frac{\pi_l}{\pi_l^{\text{ref}}}\right)\right],
\end{equation}
where $\pi_w = \pi_\theta(y_w|x)$, $\pi_l = \pi_\theta(y_l|x)$, and $\pi^{\text{ref}}$ denotes the reference policy.
We also evaluate variants including CPO and SimPO (see Section~\ref{sec:training_methods}).

\subsubsection{Explicit Reward Modeling with GRPO}

Our reward model is built by replacing the autoregressive language modeling head of an LLM with a linear layer that maps the final hidden state to a scalar reward.
We train $r_\phi$ on the preference pairs using the Bradley-Terry loss~\citep{bradley1952rank} following~\citet{ouyang2022training} and \citet{eisenstein2023helping}:
\begin{equation}
\small
\mathcal{L}_{\text{RM}} = -\mathbb{E} \left[\log \sigma(r_w - r_l)\right] + \lambda (r_w + r_l)^2,
\end{equation}
where $r_w = r_\phi(x, y_w)$, $r_l = r_\phi(x, y_l)$, and the second term encourages zero-centered rewards.

We then optimize the policy with GRPO~\citep{shao2024deepseekmath}.
For each source $x$, GRPO samples $G$ literary translations $\{y_1, \ldots, y_G\}$ from the current policy and scores them using a composite reward. The rewards are group-normalized to compute advantages:
\begin{equation}
\small
\hat{A}_i = \frac{r(x, y_i) - \text{mean}(\mathbf{r})}{\text{std}(\mathbf{r})}.
\end{equation}
The policy is optimized using a clipped objective with KL regularization:
\begin{multline}
\small
\mathcal{L}_{\text{GRPO}} = -\mathbb{E}_{x} \Bigl[ \frac{1}{G} \sum_{i=1}^G \min\bigl(\rho_i \hat{A}_i, \\
\text{clip}(\rho_i, 1{-}\epsilon, 1{+}\epsilon) \hat{A}_i\bigr) - \beta D_{\text{KL}}(\pi_\theta \| \pi_{\text{ref}}) \Bigr],
\end{multline}
where $\rho_i = \pi_\theta(y_i|x) / \pi_{\text{old}}(y_i|x)$ is the probability ratio.

\noindent\textbf{Composite Reward Design.} Following~\citet{wang2026deeptrans,wang2025extrans}, we adopt a composite reward for stable GRPO training.
However, instead of relying on expensive LLM-as-a-judge or neural network-based quality estimation (QE) models~\citep{rei2022cometkiwi}, we use a lightweight yet effective combination:
\begin{equation}
\small
r(x, y) = r_{\text{RM}} + 0.05 \cdot r_{\text{BLEU}} + r_{\text{fmt}},
\end{equation}
where $r_{\text{RM}}$ is the learned reward model score (typically in $[-3, 3]$ after training), $r_{\text{BLEU}}$ is SacreBLEU~\citep{post2018sacrebleu} (0--100 scale), and $r_{\text{fmt}}$ is a format constraint (0 if output is valid JSON with translation under the \texttt{translation} key, $-5$ otherwise). The BLEU component provides a stable lexical signal that complements the learned reward, while the format reward ensures structured outputs.

\section{Experiments}
\label{sec:experiments}

\subsection{Experimental Setup}

\paragraph{Data.} We use MetaphorTrans~\citep{wang2025drt}, an English-to-Chinese literary translation benchmark containing sentences with metaphors and similes from approximately 400 English literary works in Project Gutenberg.
The dataset comprises 19,264 training samples and 2,000 test samples.
We apply our multi-aspect pipeline to the 19,264 training sources to generate high-quality reference translations and preference pairs. For preference data, we first split at the sample level (17,337 train / 1,927 dev), then construct pairs within each split, yielding 179,588 training pairs and 19,767 dev pairs.
We also evaluate on The Essential O.\ Henry Collection~\citep{wang2026deeptrans} (586 test samples) for out-of-domain generalization.

\paragraph{Models.} We train LitMT-8B and LitMT-14B from Qwen3-8B-Base and Qwen3-14B-Base~\citep{yang2025qwen3} respectively, using our full pipeline (SFT followed by RM+GRPO).
All ablation studies use Qwen3-8B-Base.
We compare against three categories of baselines:
(1) general-purpose LLMs including Qwen3~\citep{yang2025qwen3}, Llama 3.3~\citep{grattafiori2024llama}, DeepSeek V3.1~\citep{liu2024deepseek}, Kimi K2~\citep{team2025kimi}, Claude 4.5~\citep{anthropic2025claudesonnet,anthropic2025claudeopus}, gpt-oss~\citep{agarwal2025gpt}, GPT-5.2~\citep{openai2025gpt5};
(2) general translation models such as X-ALMA~\citep{xu2024x} and LMT-60~\citep{luo2025beyond};
and (3) specialized literary translation models trained on MetaphorTrans: DRT~\citep{wang2025drt}, DeepTrans~\citep{wang2026deeptrans}, and ExTrans~\citep{wang2025extrans}.

\paragraph{Metrics.} We adopt LLM-based metrics with prompts designed for literary translation quality assessment (Appendix~\ref{app:eval_prompts}). To ensure more accurate evaluation, we upgrade the evaluator from GPT-4o used in prior work~\citep{wang2025drt,wang2026deeptrans,wang2025extrans} to Claude Opus 4.5~\citep{anthropic2025claudeopus}, one of the strongest available LLMs at the time of writing.
We report three metrics: CRF (Claude Reference-Free) and CEA5/CEA100 (Claude Evaluator Agent, 5-point and 100-point scales) following prior work~\citep{kocmi2023large,wang2025drt,wang2026deeptrans,wang2025extrans}.
CEA100 serves as the primary metric due to its finer granularity.
Unlike prior work~\citep{wang2025drt,wang2026deeptrans,wang2025extrans} that evaluates on a sampled subset, we evaluate on the full 2,000 test samples, providing more reliable and reproducible results.

\begin{table}[t]
\centering
\small
\setlength{\tabcolsep}{3pt}
\begin{tabular}{@{}llrr>{\columncolor[gray]{0.93}}r@{}}
\toprule
\textbf{Model} & \textbf{Params} & \textbf{CRF} & \textbf{CEA5} & \textbf{CEA100} \\
\midrule
X-ALMA-13B (G6) & 13B & 51.35 & 2.44 & 41.11 \\
LMT-60-8B & 8B & 54.59 & 2.59 & 44.63 \\
gpt-oss-20B$^{\spadesuit}$ & 20B (3.6B) & 55.97 & 2.59 & 44.70 \\
Llama3.3-70B & 70B & 59.46 & 2.75 & 48.34 \\
gpt-oss-120B$^{\spadesuit}$ & 120B (5.1B) & 62.25 & 2.89 & 51.50 \\
Qwen3-8B & 8B & 62.70 & 2.94 & 52.77 \\
MiniMax M2 & 230B (10B) & 65.13 & 2.96 & 53.15 \\
DRT-7B$^{\diamondsuit\spadesuit}$ & 7B & 64.48 & 3.05 & 55.09 \\
Qwen3-14B & 14B & 65.25 & 3.07 & 55.47 \\
Qwen3-32B & 32B & 66.65 & 3.15 & 57.09 \\
DRT-14B$^{\diamondsuit\spadesuit}$ & 14B & 67.07 & 3.21 & 58.43 \\
DeepTrans-7B$^{\diamondsuit\spadesuit}$ & 7B & 70.31 & 3.33 & 61.15 \\
ExTrans-7B$^{\diamondsuit\spadesuit}$ & 7B & 71.87 & 3.42 & 62.95 \\
Qwen3-235B-A22B & 235B (22B) & 74.15 & 3.55 & 65.62 \\
\rowcolor[gray]{0.92} \textbf{LitMT-8B}$^{\diamondsuit}$ & \textbf{8B} & \textbf{73.03} & \textbf{3.61} & \textbf{67.25} \\
Claude Sonnet 4.5 & -- & 77.46 & 3.66 & 68.43 \\
GPT-5.2 & -- & 77.45 & 3.68 & 68.68 \\
\rowcolor[gray]{0.92} \textbf{LitMT-14B}$^{\diamondsuit}$ & \textbf{14B} & \textbf{75.20} & \textbf{3.71} & \textbf{69.07} \\
DeepSeek V3.1 & 671B (37B) & 77.19 & 3.75 & 70.27 \\
Kimi K2$^{\spadesuit}$ & 1T (32B) & 79.93 & 3.81 & 71.70 \\
Claude Opus 4.5 & -- & 79.95 & 3.90 & 73.30 \\
\midrule
\multicolumn{5}{l}{\scriptsize $^{\diamondsuit}$trained on MetaphorTrans. $^{\spadesuit}$thinking-enabled. \textbf{Bold}/\colorbox[gray]{0.92}{gray}: our models.} \\
\bottomrule
\end{tabular}
\caption{Main results on MetaphorTrans (in-domain). Models sorted by CEA100. Params in parentheses denote activated parameters for MoE models. Qwen3-235B-A22B refers to the Instruct variant. gpt-oss models use medium reasoning effort.}
\label{tab:main_results}
\end{table}

\begin{table}[t]
\centering
\small
\setlength{\tabcolsep}{3pt}
\begin{tabular}{@{}llrr>{\columncolor[gray]{0.93}}r@{}}
\toprule
\textbf{Model} & \textbf{Params} & \textbf{CRF} & \textbf{CEA5} & \textbf{CEA100} \\
\midrule
X-ALMA-13B (G6) & 13B & 59.05 & 2.78 & 48.59 \\
gpt-oss-20B$^{\spadesuit}$ & 20B (3.6B) & 63.68 & 2.91 & 51.63 \\
LMT-60-8B & 8B & 66.78 & 3.15 & 56.88 \\
Llama3.3-70B & 70B & 68.57 & 3.18 & 57.96 \\
Qwen3-8B & 8B & 70.20 & 3.30 & 60.56 \\
DRT-7B$^{\diamondsuit\spadesuit}$ & 7B & 70.30 & 3.31 & 60.67 \\
gpt-oss-120B$^{\spadesuit}$ & 120B (5.1B) & 71.35 & 3.34 & 61.49 \\
MiniMax M2 & 230B (10B) & 73.49 & 3.35 & 61.79 \\
DeepTrans-7B$^{\diamondsuit\spadesuit}$ & 7B & 74.58 & 3.46 & 64.17 \\
Qwen3-14B & 14B & 73.38 & 3.49 & 64.32 \\
DRT-14B$^{\diamondsuit\spadesuit}$ & 14B & 73.67 & 3.52 & 65.19 \\
Qwen3-32B & 32B & 75.51 & 3.56 & 65.81 \\
ExTrans-7B$^{\diamondsuit\spadesuit}$ & 7B & 75.04 & 3.62 & 67.39 \\
\rowcolor[gray]{0.92} \textbf{LitMT-8B}$^{\diamondsuit}$ & \textbf{8B} & \textbf{76.42} & \textbf{3.77} & \textbf{70.38} \\
\rowcolor[gray]{0.92} \textbf{LitMT-14B}$^{\diamondsuit}$ & \textbf{14B} & \textbf{79.65} & \textbf{3.94} & \textbf{73.71} \\
Qwen3-235B-A22B & 235B (22B) & 81.61 & 3.94 & 74.01 \\
Claude Sonnet 4.5 & -- & 84.27 & 4.05 & 76.50 \\
GPT-5.2 & -- & 84.47 & 4.12 & 77.74 \\
DeepSeek V3.1 & 671B (37B) & 83.56 & 4.12 & 77.94 \\
Kimi K2$^{\spadesuit}$ & 1T (32B) & 85.55 & 4.16 & 78.56 \\
Claude Opus 4.5 & -- & 86.22 & 4.26 & 80.65 \\
\midrule
\multicolumn{5}{l}{\scriptsize $^{\diamondsuit}$trained on MetaphorTrans. $^{\spadesuit}$thinking-enabled. \textbf{Bold}/\colorbox[gray]{0.92}{gray}: our models.} \\
\bottomrule
\end{tabular}
\caption{Out-of-domain results on O.\ Henry Collection. Models sorted by CEA100. Qwen3-235B-A22B refers to the Instruct variant. gpt-oss models use medium reasoning effort.}
\label{tab:ohenry_results}
\end{table}

\paragraph{Implementation.} Our refinement pipeline uses Qwen3-235B-A22B-Instruct~\citep{yang2025qwen3,qwen2025qwen3instruct} as the backbone for all modules, with maximum rounds $K=8$, early stopping patience $N=3$, and score threshold $\tau=4.9$ (out of 5.0). All training experiments use 8$\times$H100 GPUs with global batch size 128. For SFT and DPO-series methods: learning rate 1e-5, warmup ratio 0.05, 3 epochs. The reward model trains for 1 epoch with $\lambda=0.01$ from the SFT checkpoint. GRPO uses $G=16$ generations per sample, temperature 1.0, top-p 0.9, $\beta=0.01$, learning rate 1e-7, 3 epochs from the SFT checkpoint.

\begin{figure}[t]
\centering
\includegraphics[width=\columnwidth]{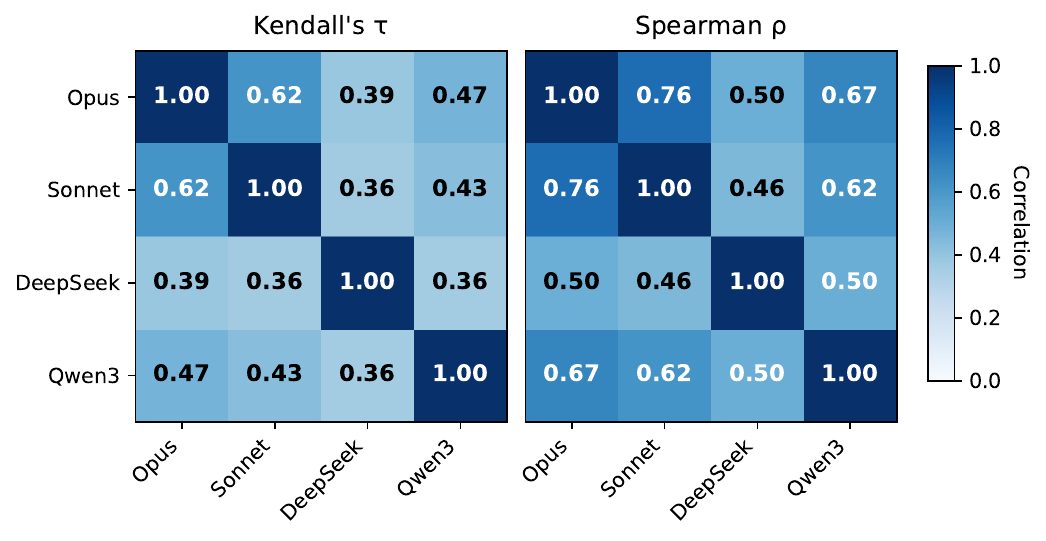}
\caption{Sample-level correlation between four LLM judges over 12,000 instances. Left: Kendall's $\tau$. Right: Spearman's $\rho$. All correlations significant at $p < 0.001$.}
\label{fig:judge_correlation}
\end{figure}

\subsection{Main Results}

Table~\ref{tab:main_results} presents results on MetaphorTrans.
Our models demonstrate strong parameter efficiency: LitMT-8B (8B) outperforms Qwen3-235B-A22B (235B, 22B activated) by 1.63 CEA100 points, achieving comparable quality using 30$\times$ fewer parameters.
Notably, our training data is generated using Qwen3-235B-A22B, yet the 8B student model surpasses its teacher---demonstrating that our multi-aspect refinement pipeline genuinely improves data quality rather than merely distilling knowledge.
LitMT-14B achieves 69.07 CEA100, competitive with frontier models such as Claude Sonnet 4.5 (68.43) and GPT-5.2 (68.68).
Table~\ref{tab:ohenry_results} presents out-of-domain results on The Essential O.\ Henry Collection, which features distinct narrative styles and early 20th-century American English. Our models demonstrate strong generalization: LitMT-14B (73.71) approaches Qwen3-235B-A22B (74.01), and LitMT-8B (70.38) surpasses Qwen3-32B (65.81) by a wide margin.

\subsection{Evaluation Reliability}
\label{sec:eval_reliability}

Human evaluation of literary translation requires bilingual proficiency and literary expertise to assess stylistic nuances---such expertise is scarce and costly at scale. To nevertheless validate our LLM-based evaluation, we conduct inter-judge agreement analysis following recent advances~\citep{kocmi2023large,zheng2023llmjudge}.

We assess agreement using four frontier LLMs: Claude Opus 4.5, Claude Sonnet 4.5, DeepSeek V3.1, and Qwen3-235B-A22B-Instruct. We select six representative models spanning different performance levels and compute pairwise system rankings, yielding $\binom{6}{2}=15$ comparisons per judge. The judges achieve 88.9\% agreement on relative ranking at the system level (see Appendix~\ref{app:judge_agreement} for details). At the sample level, Figure~\ref{fig:judge_correlation} shows correlations over 12,000 instances, with Kendall's $\tau$ ranging from 0.36 to 0.62---comparable to or exceeding the $\tau$ of 0.36--0.38 reported between GPT-4 and human~\citep{kocmi2023large}. Cross-family judge pairs also exhibit substantial agreement (e.g., Claude--Qwen $\tau$=0.47), confirming that our evaluation captures genuine quality distinctions.

\subsection{Ablation Studies}

\subsubsection{Effect of Training Methods}
\label{sec:training_methods}

Table~\ref{tab:training_methods} compares training methods using identical data. All preference optimization methods (DPO-series and RM+GRPO) are initialized from the SFT checkpoint. A notable finding is that all DPO-series methods underperform SFT, with performance drops of 2--3 CEA100 points. In contrast, RM+GRPO improves over SFT by 1.51 points, validating our hypothesis that explicit reward modeling with online exploration better leverages the preference data.

\begin{table}[t]
\centering
\small
\setlength{\tabcolsep}{4pt}
\begin{tabular}{@{}lrr>{\columncolor[gray]{0.93}}r@{}}
\toprule
\textbf{Method} & \textbf{CRF} & \textbf{CEA5} & \textbf{CEA100} \\
\midrule
SFT & \underline{72.66} & \underline{3.54} & \underline{65.74} \\
\midrule
SimPO & 69.98 & 3.41 & 62.62 \\
DPO & 70.50 & 3.44 & 63.39 \\
CPO & 70.69 & 3.45 & 63.67 \\
\midrule
\rowcolor[gray]{0.92} \textbf{RM+GRPO} & \textbf{73.03} & \textbf{3.61} & \textbf{67.25} \\
\bottomrule
\end{tabular}
\caption{Effect of training methods. Explicit reward modeling with GRPO outperforms all implicit reward methods.}
\label{tab:training_methods}
\end{table}

\subsubsection{Effect of Multi-Aspect Refinement}

Table~\ref{tab:pipeline_ablation} examines each component's contribution.
Best-of-8~\citep{stiennon2020learning} generates 8 translations via temperature sampling and selects the best; this baseline performs worst, confirming that targeted refinement outperforms random exploration.
Naive-only uses standard iterative refinement without specialized modules.
Among single-agent variants, Literary Preserver-only achieves the strongest results (66.91), suggesting that literary effect preservation is more challenging to optimize.
Multi-aspect refinement combines both agents' strengths, achieving the best overall performance (67.25).
\begin{table}[t]
\centering
\small
\setlength{\tabcolsep}{4pt}
\begin{tabular}{@{}lrr>{\columncolor[gray]{0.93}}r@{}}
\toprule
\textbf{Variant} & \textbf{CRF} & \textbf{CEA5} & \textbf{CEA100} \\
\midrule
Best-of-8 Sampling & 71.44 & 3.41 & 63.03 \\
\midrule
Naive-only & 72.92 & 3.54 & 65.64 \\
Expression Optimizer-only & 72.21 & 3.52 & 65.27 \\
Literary Preserver-only & \textbf{73.39} & \underline{3.60} & \underline{66.91} \\
\midrule
\rowcolor[gray]{0.92} \textbf{All (Multi-Aspect)} & \underline{73.03} & \textbf{3.61} & \textbf{67.25} \\
\bottomrule
\end{tabular}
\caption{Effect of multi-aspect refinement.}
\label{tab:pipeline_ablation}
\end{table}

\subsubsection{Effect of Data Source}

Table~\ref{tab:data_quality} compares SFT performance across different training data sources. Our pipeline uses Qwen3-235B-A22B as the backbone for data generation. Compared to single-pass distillation from the same model, our multi-aspect refinement improves CEA100 by 4.66 points (61.08 $\rightarrow$ 65.74), demonstrating that iterative refinement genuinely enhances data quality beyond simple distillation. Our data also outperforms the original DRT ground truth by 8.65 points.

\begin{table}[t]
\centering
\small
\setlength{\tabcolsep}{4pt}
\begin{tabular}{@{}lrr>{\columncolor[gray]{0.93}}r@{}}
\toprule
\textbf{Data Source} & \textbf{CRF} & \textbf{CEA5} & \textbf{CEA100} \\
\midrule
DRT Ground Truth & 66.41 & 3.15 & 57.09 \\
Qwen3-235B-A22B Distill & 70.02 & 3.32 & 61.08 \\
DeepSeek V3.1 Distill & \underline{71.04} & \underline{3.40} & \underline{62.78} \\
\midrule
\rowcolor[gray]{0.92} \textbf{Ours} & \textbf{72.66} & \textbf{3.54} & \textbf{65.74} \\
\bottomrule
\end{tabular}
\caption{Effect of data source (SFT only).}
\label{tab:data_quality}
\end{table}

\subsubsection{Effect of Base Model}
\label{sec:base_model}

Table~\ref{tab:base_model} examines the effect of base model choice. Qwen3-8B-Base achieves the best performance (67.25 CEA100), outperforming Qwen3-8B by 1.48 points. However, for Qwen2.5-7B, the Instruct variant slightly outperforms the base model (64.40 vs.\ 63.97). The most significant factor is model family: Qwen3 models consistently outperform Qwen2.5 models by 2.8--3.3 points regardless of variant.

\begin{table}[t]
\centering
\small
\setlength{\tabcolsep}{4pt}
\begin{tabular}{@{}lrr>{\columncolor[gray]{0.93}}r@{}}
\toprule
\textbf{Base Model} & \textbf{CRF} & \textbf{CEA5} & \textbf{CEA100} \\
\midrule
Qwen2.5-7B & 71.05 & 3.47 & 63.97 \\
Qwen2.5-7B-Instruct & 71.46 & 3.48 & 64.40 \\
Qwen3-8B & \underline{72.12} & \underline{3.55} & \underline{65.77} \\
\midrule
\rowcolor[gray]{0.92} \textbf{Qwen3-8B-Base} & \textbf{73.03} & \textbf{3.61} & \textbf{67.25} \\
\bottomrule
\end{tabular}
\caption{Effect of base model choice with our training approach.}
\label{tab:base_model}
\end{table}

\subsubsection{Data Generation Statistics}
\label{sec:data_gen_stats}

Table~\ref{tab:data_gen_stats} presents statistics from our multi-aspect iterative refinement pipeline. The pipeline achieves an average score improvement of 0.31 points (from 4.43 to 4.73 on a 5-point scale), with 61.6\% of samples reaching the quality threshold ($\tau=4.9$). The average best-worst score difference of 0.70 indicates meaningful quality distinctions within each sample's iteration history, providing informative preference pairs for training.

\begin{table}[t]
\centering
\small
\setlength{\tabcolsep}{6pt}
\begin{tabular}{@{}lr@{}}
\toprule
\textbf{Statistic} & \textbf{Value} \\
\midrule
Total samples & 19,264 \\
Average refinement rounds & 4.08 \\
\midrule
Average initial score & 4.43 \\
Average final score & 4.73 \\
Average best score & 4.88 \\
Average worst score & 4.19 \\
\midrule
Average score improvement & 0.31 \\
Average best-worst difference & 0.70 \\
Samples reached threshold & 11,863 (61.6\%) \\
\bottomrule
\end{tabular}
\caption{Data generation statistics from the multi-aspect iterative refinement pipeline on MetaphorTrans training set.}
\label{tab:data_gen_stats}
\end{table}

\subsubsection{Reward Model Performance}
\label{sec:rm_performance}

Table~\ref{tab:rm_accuracy} reports the pairwise classification accuracy of our 8B reward model (\S\ref{sec:training}) on the held-out RM Dev set comprising 19,767 preference pairs. Results are stratified by the absolute score margin between chosen and rejected translations. The model attains 72.49\% overall accuracy, with accuracy ranging from around 68\% on ambiguous pairs (margin $<$0.5) to 96.20\% on unambiguous cases (margin $\geq$3.0). This calibration validates that our preference data encodes learnable quality signals spanning the full difficulty spectrum.

\begin{table}[t]
\centering
\small
\setlength{\tabcolsep}{4pt}
\begin{tabular}{@{}lrr>{\columncolor[gray]{0.93}}r@{}}
\toprule
\textbf{Score Diff} & \textbf{Total} & \textbf{Correct} & \textbf{Acc.\ (\%)} \\
\midrule
$[0, 0.25)$ & 10,845 & 7,473 & 68.91 \\
$[0.25, 0.5)$ & 2,336 & 1,583 & 67.77 \\
$[0.5, 1.0)$ & 2,287 & 1,695 & 74.11 \\
$[1.0, 1.5)$ & 2,170 & 1,761 & 81.15 \\
$[1.5, 2.0)$ & 745 & 587 & 78.79 \\
$[2.0, 2.5)$ & 946 & 813 & 85.94 \\
$[2.5, 3.0)$ & 201 & 189 & 94.03 \\
$\geq 3.0$ & 237 & 228 & 96.20 \\
\midrule
\textbf{Overall} & \textbf{19,767} & \textbf{14,329} & \textbf{72.49} \\
\bottomrule
\end{tabular}
\caption{Reward model (8B) accuracy on RM Dev set by score difference (0--5 scale) between chosen and rejected translations.}
\label{tab:rm_accuracy}
\end{table}

\section{Conclusion}

We present a multi-aspect data generation and LLM training approach for literary translation. By decomposing quality into expression fluency and literary effect, our pipeline generates high-quality references and preference pairs. The generated data outperforms existing ground truth for SFT by 8.65 CEA100 points. For preference optimization, explicit reward modeling with GRPO yields an additional 1.51 points, while DPO-series methods degrade performance. Our models, LitMT-8B and LitMT-14B, achieve 67.25 and 69.07 CEA100 on MetaphorTrans, competitive with Claude Sonnet 4.5 (68.43), and demonstrate strong generalization on out-of-domain O.\ Henry data. Our approach provides a strong baseline for future literary translation research.

\section*{Limitations}

Our work has several limitations. First, while we empirically show that explicit reward modeling with GRPO outperforms DPO-series methods, a theoretical understanding of this phenomenon remains an open question.
Second, whether incorporating long chain-of-thought reasoning~\citep{wang2025drt,wang2026deeptrans,wang2025extrans} into our framework yields further gains remains to be explored.
Third, our experiments focus on English-to-Chinese; extending to other language pairs is left for future work.

\cleardoublepage

\section*{Ethical Considerations}

We discuss the main ethical considerations of our work as follows:
(1) \textbf{Copyright.} We use the MetaphorTrans benchmark~\citep{wang2025drt}, which mines literary sentences from approximately 400 English books provided by the Project Gutenberg public-domain book repository.\footnote{\url{https://www.gutenberg.org/}} These books are typically more than fifty years old and their copyrights have expired. The O.\ Henry Collection used for out-of-domain evaluation is also in the public domain.
(2) \textbf{Potential Biases.} The backbone models used in our data generation pipeline (Qwen3-235B-A22B-Instruct) and training (Qwen3-8B-Base, Qwen3-14B-Base) may exhibit biases present in their pretraining data. Our models, LitMT-8B and LitMT-14B, might inherit similar biases.


\bibliography{custom}

\cleardoublepage
\newpage

\appendix

\section{Additional Experiments}
\label{app:additional_exp}

\subsection{Effect of Composite Reward Design}
\label{app:reward_ablation}

Table~\ref{tab:reward_ablation} presents an ablation study on the reward weighting coefficients in our composite reward formulation $r = w_{\text{RM}} \cdot r_{\text{RM}} + w_{\text{BLEU}} \cdot r_{\text{BLEU}} + r_{\text{fmt}}$ (\S\ref{sec:training}). Employing the learned reward in isolation ($w_{\text{BLEU}}=0$) yields 64.48 CEA100. Incorporating BLEU as a complementary lexical signal progressively improves performance, with optimal results at $w_{\text{BLEU}}=0.05$ (67.25). However, increasing the BLEU coefficient to 0.1 degrades performance to 66.80, indicating that excessive reliance on surface-level lexical matching is detrimental. Conversely, using BLEU alone ($w_{\text{RM}}=0$) achieves 65.49, demonstrating that the learned reward model captures quality dimensions beyond n-gram overlap. These results suggest that the optimal reward design balances learned quality judgments with lexical grounding.

\begin{table}[h]
\centering
\small
\setlength{\tabcolsep}{4pt}
\begin{tabular}{@{}cc rr>{\columncolor[gray]{0.93}}r@{}}
\toprule
$w_{\text{RM}}$ & $w_{\text{BLEU}}$ & \textbf{CRF} & \textbf{CEA5} & \textbf{CEA100} \\
\midrule
1 & 0 & 69.71 & 3.50 & 64.48 \\
1 & 0.01 & 71.86 & 3.58 & 66.10 \\
\rowcolor[gray]{0.92} \textbf{1} & \textbf{0.05} & 73.11 & 3.60 & \textbf{67.25} \\
1 & 0.1 & 73.40 & 3.59 & 66.80 \\
\hdashline
0 & 1 & 72.75 & 3.53 & 65.49 \\
\bottomrule
\end{tabular}
\caption{Effect of reward weighting in composite reward design. Performance peaks at $w_{\text{BLEU}}=0.05$; higher weights degrade quality.}
\label{tab:reward_ablation}
\end{table}

\section{Evaluation Reliability Analysis}
\label{app:judge_agreement}

\begin{table}[t]
\centering
\small
\setlength{\tabcolsep}{3pt}
\begin{tabular}{@{}lllll@{}}
\toprule
\textbf{Model} & \textbf{Opus} & \textbf{Sonnet} & \textbf{DeepSeek} & \textbf{Qwen3} \\
\midrule
Kimi K2 & 71.70 (1) & 78.06 (1) & 87.19 (4) & 82.01 (2) \\
\rowcolor[gray]{0.92} LitMT-14B & 69.07 (2) & 74.50 (2) & 88.51 (1) & 82.29 (1) \\
\rowcolor[gray]{0.92} LitMT-8B & 67.25 (3) & 71.94 (3) & 88.09 (2) & 81.70 (3) \\
DeepTrans-7B & 61.15 (4) & 64.41 (4) & 88.07 (3) & 79.30 (4) \\
DRT-7B & 55.09 (5) & 62.07 (5) & 81.22 (5) & 67.99 (5) \\
Qwen3-8B & 52.77 (6) & 59.96 (6) & 77.17 (6) & 64.22 (6) \\
\bottomrule
\end{tabular}
\caption{CEA100 scores (rank) from four LLM judges on MetaphorTrans test set. Opus = Claude Opus 4.5, Sonnet = Claude Sonnet 4.5, DeepSeek = DeepSeek V3.1, Qwen3 = Qwen3-235B-A22B-Instruct. Models sorted by Opus score.}
\label{tab:judge_scores}
\end{table}

This section provides detailed analysis supporting the evaluation reliability results presented in \S\ref{sec:eval_reliability}.

Human evaluation of literary translation poses distinctive methodological challenges: annotators must demonstrate bilingual proficiency alongside sufficient literary expertise to assess stylistic nuances, figurative language preservation, and aesthetic quality. Such specialized expertise is scarce and prohibitively expensive to obtain at scale. Following recent advances in LLM-based evaluation~\citep{kocmi2023large,zheng2023llmjudge}, we adopt automated assessment and conduct a comprehensive reliability analysis to validate our evaluation protocol.

\paragraph{Model-Level Consistency.}
We evaluate six representative systems on the MetaphorTrans test set using four frontier LLMs as judges: Claude Opus 4.5~\citep{anthropic2025claudeopus}, Claude Sonnet 4.5~\citep{anthropic2025claudesonnet}, DeepSeek V3.1~\citep{liu2024deepseek}, and Qwen3-235B-A22B-Instruct~\citep{yang2025qwen3,qwen2025qwen3instruct}. Table~\ref{tab:judge_scores} reports CEA100 scores with corresponding rankings from each judge. Despite variation in absolute score magnitudes, all four judges consistently rank LitMT-8B and LitMT-14B above the baseline systems (DRT-7B, DeepTrans-7B, Qwen3-8B). Across all 15 pairwise system comparisons, the judges exhibit 88.9\% agreement on relative ordering, providing robust evidence that our principal findings are judge-invariant.

\paragraph{Sample-Level Correlation.}
We compute pairwise correlations between all judge combinations over 12,000 evaluation instances (2,000 test samples $\times$ 6 systems). Figure~\ref{fig:judge_correlation} (in \S\ref{sec:eval_reliability}) visualizes Kendall's $\tau$ and Spearman's $\rho$ coefficients. All correlations achieve statistical significance ($p < 0.001$). The inter-judge Kendall's $\tau$ ranges from 0.36 to 0.62, comparable to or exceeding the $\tau$ of 0.36--0.38 reported between GPT-4 and human MQM annotations~\citep{kocmi2023large}. This indicates that our LLM judges exhibit mutual agreement at levels commensurate with human--LLM concordance. Notably, cross-family judge pairs also demonstrate substantial correlation (e.g., Claude--Qwen $\tau$=0.47, Claude--DeepSeek $\tau$=0.39), confirming that our evaluation captures genuine quality distinctions rather than model-specific scoring artifacts.

\paragraph{Evaluation Stability.}
To assess reproducibility, we execute CEA100 evaluation with Claude Opus 4.5 on LitMT-8B across three independent runs. The resulting scores are 67.25, 67.25, and 67.23, yielding a standard deviation of 0.01. This negligible variance demonstrates that our LLM-based evaluation protocol produces highly stable and reproducible measurements.

\begin{table*}[t]
\centering
\small
\setlength{\tabcolsep}{3pt}
\begin{tabular}{@{}p{2.5cm}p{13cm}@{}}
\toprule
\textbf{Source} & This it was, more than any thing else, that roused such a \textbf{tempest} in my \textbf{soul}. \\
\midrule
Qwen3-8B & \foreignlanguage{chinese}{正是这一点，比任何其他事情都更激起了我\textcolor{red}{内心}的巨大风暴。} \\
DRT-7B & \foreignlanguage{chinese}{\textcolor{red}{这}比任何事情都更\textcolor{red}{深深}激起了我内心中的这场风暴。} \\
DeepTrans-7B & \foreignlanguage{chinese}{正是这一点，胜过一切其他因素，在我\textcolor{red}{心中}激起了如此\textcolor{green!60!black}{狂风巨浪}。} \\
\textbf{LitMT-8B} & \foreignlanguage{chinese}{正是这一点，胜过其他任何事，激起了我\textcolor{green!60!black}{灵魂}中的\textcolor{green!60!black}{狂风骤雨}。} \\
\textbf{LitMT-14B} & \foreignlanguage{chinese}{正是这一点，胜过世间一切，于我\textcolor{green!60!black}{心魂}深处掀起\textcolor{green!60!black}{滔天巨浪}。} \\
\midrule
\textbf{Source} & O, then the earth shook to see the \textbf{heavens on fire}, And not in fear of your \textbf{nativity}. \\
\midrule
Qwen3-8B & \foreignlanguage{chinese}{哦，大地震动，目睹苍穹燃起\textcolor{red}{战火}，却并非因你\textcolor{red}{生辰的恐惧}。} \\
DRT-7B & \foreignlanguage{chinese}{哦，于是大地因\textcolor{green!60!black}{天火熊熊}而震颤，非因惧您\textcolor{green!60!black}{降临尘世}。} \\
DeepTrans-7B & \foreignlanguage{chinese}{哦，于是大地震颤，目睹苍穹燃起\textcolor{green!60!black}{熊熊烈焰}，却非因你\textcolor{green!60!black}{降世}而生畏惧。} \\
\textbf{LitMT-8B} & \foreignlanguage{chinese}{啊，大地因\textcolor{green!60!black}{苍天燃起烈火}而震颤，却并非畏惧你\textcolor{green!60!black}{降生}的威仪。} \\
\textbf{LitMT-14B} & \foreignlanguage{chinese}{啊，大地震颤，只因\textcolor{green!60!black}{天穹烈焰腾起}，并非因惧你\textcolor{green!60!black}{降生}而\textcolor{green!60!black}{惶然惊悸}。} \\
\midrule
\textbf{Source} & He said he'd \textbf{cowhide} me till I was \textbf{black and blue} if I didn't raise some money for him. \\
\midrule
Qwen3-8B & \foreignlanguage{chinese}{他说如果我不为他筹些钱，他就会把我\textcolor{red}{打}得\textcolor{green!60!black}{青一块紫一块}。} \\
DRT-7B & \foreignlanguage{chinese}{他说，若我不帮他弄到些钱，他就会用\textcolor{green!60!black}{牛皮鞭子}狠狠地抽我，直到我\textcolor{red}{遍体鳞伤}。} \\
DeepTrans-7B & \foreignlanguage{chinese}{他扬言要\textcolor{red}{重重抽我一顿}，直到我浑身\textcolor{green!60!black}{青一块紫一块}，除非我给他筹来些钱。} \\
\textbf{LitMT-8B} & \foreignlanguage{chinese}{他说，如果我不为他弄到钱，就用\textcolor{green!60!black}{牛皮}抽我，直到我\textcolor{green!60!black}{青一块紫一块}。} \\
\textbf{LitMT-14B} & \foreignlanguage{chinese}{他说，如果我不替他弄到钱，就拿\textcolor{green!60!black}{牛皮鞭}把我抽得\textcolor{green!60!black}{青一块紫一块}。} \\
\bottomrule
\end{tabular}
\caption{Case studies of literary translation. \textcolor{green!60!black}{Green} indicates good word choices, while \textcolor{red}{red} indicates errors. Example 1: ``soul'' should be \foreignlanguage{chinese}{灵魂/心魂} not \foreignlanguage{chinese}{内心/心中}; ``tempest'' rendered as \foreignlanguage{chinese}{狂风骤雨/滔天巨浪} is more vivid. Example 2: ``heavens on fire'' should not add ``war'' (\foreignlanguage{chinese}{战火}); ``nativity'' means birth (\foreignlanguage{chinese}{降生}), not birthday (\foreignlanguage{chinese}{生辰}). Example 3: ``cowhide'' specifically means whipping with leather (\foreignlanguage{chinese}{牛皮/牛皮鞭}); ``black and blue'' means bruises (\foreignlanguage{chinese}{青一块紫一块}), not wounds (\foreignlanguage{chinese}{遍体鳞伤}).}
\label{tab:case_study}
\end{table*}

\section{Translation Case Studies}
\label{app:case_studies}

Table~\ref{tab:case_study} presents representative examples comparing translations from different models. \textcolor{green!60!black}{Green} indicates good word choices, while \textcolor{red}{red} indicates problematic translations.

\section{Cost Analysis}
\label{app:cost_analysis}

We compare the computational costs of our approach against methods that use LLM-as-a-Judge for online reward signals during GRPO training. All experiments use the MetaphorTrans training set (19,264 samples).

\subsection{LLM API Costs}
\label{app:api_costs}

Table~\ref{tab:api_costs} presents the token consumption and associated costs. We estimate token usage as follows:

\paragraph{Our Method.} The multi-aspect iterative refinement pipeline (\S\ref{sec:refinement}) processes each sample through naive translation, expression optimization, literary effect preservation, aggregation, and evaluation, with an average of 4.08 refinement rounds (Table~\ref{tab:data_gen_stats}). This is a \textbf{one-time} data generation cost---the resulting SFT data and preference pairs can be reused across multiple training runs.

\paragraph{DeepTrans/ExTrans.} These methods require LLM evaluation during GRPO training. We estimate costs for a single training run (2 epochs, 8 rollouts per sample). DeepTrans~\citep{wang2026deeptrans} uses DeepSeek-V3 for scoring; ExTrans~\citep{wang2025extrans} uses DeepSeek-R1 to generate reference translations and DeepSeek-V3 to compare model outputs against these references.

\begin{table}[t]
\centering
\small
\setlength{\tabcolsep}{3pt}
\begin{tabular}{@{}llrrr@{}}
\toprule
\textbf{Method} & \textbf{Model} & \textbf{Input} & \textbf{Output} & \textbf{Cost} \\
\midrule
\rowcolor[gray]{0.92} \textbf{Ours} & Qwen3-235B & 192.8M & 143.2M & \textbf{\$168} \\
\midrule
DeepTrans & DeepSeek-V3 & 122.2M & 146.8M & \$318 \\
\midrule
\multirow{2}{*}{ExTrans} & DeepSeek-R1 & 3.8M & 175.9M & \$955 \\
 & DeepSeek-V3 & 116.7M & 9.2M & \$83 \\
\cmidrule{2-5}
 & \textit{Total} & -- & -- & \$1,038 \\
\bottomrule
\end{tabular}
\caption{LLM API costs. Our method is one-time data generation; DeepTrans/ExTrans costs are per GRPO run (2 epochs, 8 rollouts). Prices from AWS Bedrock~\citep{aws2025bedrock}: Qwen3-235B (\$0.22/\$0.88 per 1M input/output tokens), DeepSeek-V3.1 (\$0.58/\$1.68, as proxy for V3), DeepSeek-R1 (\$1.35/\$5.40).}
\label{tab:api_costs}
\end{table}

Since DeepTrans and ExTrans compute reward signals online, they require fresh API calls for every training run. For a typical development cycle with 5--10 hyperparameter iterations, our cost remains \$168 while DeepTrans scales to \$1,590--\$3,180 and ExTrans to \$5,190--\$10,380.

\subsection{GPU Training Costs}
\label{app:gpu_costs}

Training LitMT-8B requires 80 H100 GPU hours: 2 hours for SFT, 8 hours for reward model training, and 70 hours for GRPO. The reward model enables training without any LLM API calls. Adopting online LLM-as-a-Judge as in DeepTrans or ExTrans would increase GPU costs by several times or even an order of magnitude, as GPUs idle while waiting for API responses.

\section{Data Generation Prompts}
\label{app:data_gen_prompts}

The following prompts are used in our multi-aspect iterative refinement pipeline (\S\ref{sec:refinement}):
\begin{itemize}
    \item Naive Translator (Appendix~\ref{app:naive_translator})
    \item Expression Optimizer (Appendix~\ref{app:expression_optimizer})
    \item Literary Effect Preserver (Appendix~\ref{app:literary_preserver})
    \item Aggregator (Appendix~\ref{app:aggregator})
    \item Evaluator (Appendix~\ref{app:evaluator})
\end{itemize}

\begin{figure*}[t]
\subsection{Naive Translator}
\label{app:naive_translator}
\begin{tcblisting}{colback=gray!8, colframe=gray!40, listing only, listing options={basicstyle=\footnotesize\ttfamily, breaklines=true, breakatwhitespace=false, columns=flexible}}
Please translate the following text from English to {target_language}. Output your translation in XML format.

Input text:
{source}

Output format:
<translation>
your translation here
</translation>
\end{tcblisting}
\end{figure*}

\begin{figure*}[t]
\subsection{Expression Optimizer}
\label{app:expression_optimizer}
\begin{tcblisting}{colback=gray!8, colframe=gray!40, listing only, listing options={basicstyle=\footnotesize\ttfamily, breaklines=true, breakatwhitespace=false, columns=flexible}}
You are an expression optimization expert. Your role is to ensure the translation reads naturally and fluently in {target_language}, eliminating translationese and awkward phrasing.

<source_text>
{source}
</source_text>

<current_translation>
{translation}
</current_translation>

<evaluator_feedback>
{evaluator_feedback}
</evaluator_feedback>

Analyze the translation for expression quality issues:

1. Word Order: Does it follow natural {target_language} syntax?
   - Subject-verb-object arrangement
   - Modifier placement (adjectives, adverbs)
   - Clause ordering and sentence flow

2. Collocation: Are word combinations natural and idiomatic?
   - Verb-noun, adjective-noun pairings
   - Common {target_language} expressions vs. literal translations
   - Avoiding awkward or non-native combinations

3. Redundancy: Is the expression concise without unnecessary repetition?
   - Removing verbose constructions
   - Eliminating translationese patterns
   - Maintaining clarity while being concise

Provide specific improvements to enhance naturalness and fluency.

Output in XML format:
<result>
<issues>
Specific expression issues: word order, collocation, or redundancy problems
</issues>
<improved_translation>
Improved translation (return original if no improvement needed)
</improved_translation>
</result>
\end{tcblisting}
\end{figure*}

\begin{figure*}[t]
\subsection{Literary Effect Preserver}
\label{app:literary_preserver}
\begin{tcblisting}{colback=gray!8, colframe=gray!40, listing only, listing options={basicstyle=\footnotesize\ttfamily, breaklines=true, breakatwhitespace=false, columns=flexible}}
You are a literary effect expert. Your role is to preserve and recreate the literary and aesthetic qualities of the source text, including metaphors, rhetorical devices, tone, and emotional impact.

<source_text>
{source}
</source_text>

<current_translation>
{translation}
</current_translation>

<evaluator_feedback>
{evaluator_feedback}
</evaluator_feedback>

Analyze the translation for literary quality:

1. Metaphors and Imagery: Are figurative expressions appropriately handled?
   - Identify metaphors, similes, and imagery in the source
   - Decide: preserve literally, adapt culturally, or explicitate?
   - Example: "drowning in work" can be preserved literally or adapted idiomatically
   - Ensure metaphors are understandable in target culture

2. Rhetorical Devices: Are stylistic features recreated?
   - Parallelism, repetition, alliteration, rhythm
   - Wordplay, puns, double meanings
   - Can these be recreated or compensated in {target_language}?

3. Tone and Emotional Impact: Does it evoke similar feelings?
   - Formal/informal, serious/humorous, poetic/prosaic
   - Emotional atmosphere and mood
   - Cultural connotations and associations

Provide specific improvements to enhance literary and aesthetic quality.

Output in XML format:
<result>
<issues>
Specific literary issues: metaphors, rhetorical devices, or tone/emotion problems
</issues>
<improved_translation>
Improved translation (return original if no improvement needed)
</improved_translation>
</result>
\end{tcblisting}
\end{figure*}

\begin{figure*}[t]
\subsection{Aggregator}
\label{app:aggregator}
\begin{tcblisting}{colback=gray!8, colframe=gray!40, listing only, listing options={basicstyle=\footnotesize\ttfamily, breaklines=true, breakatwhitespace=false, columns=flexible}}
You are a translation integration expert. You have translation results from two experts. Please synthesize this information and provide an optimal translation.

<source_text>
{source}
</source_text>

<expression_optimization_expert>
{expression_translation}
</expression_optimization_expert>

<literary_effect_expert>
{literary_translation}
</literary_effect_expert>

Please synthesize the two experts' translations and provide an optimal translation that balances expression fluency and literary effect.

Output in XML format:
<result>
<reasoning>
Integration reasoning explanation
</reasoning>
<final_translation>
Final translation
</final_translation>
</result>
\end{tcblisting}
\end{figure*}

\begin{figure*}[t]
\subsection{Evaluator}
\label{app:evaluator}
\begin{tcblisting}{colback=gray!8, colframe=gray!40, listing only, listing options={basicstyle=\footnotesize\ttfamily, breaklines=true, breakatwhitespace=false, columns=flexible}}
[System]
Please evaluate the following {target_language} translation of an English text. Rate the translation on a scale of 0.00 to 5.00 (with two decimal places), where:
- 1: Poor translation; the text is somewhat understandable but contains significant errors and awkward phrasing that greatly hinder comprehension for a {target_language} reader.
- 2: Fair translation; the text conveys the basic meaning but lacks fluency and contains several awkward phrases or inaccuracies, making it challenging for a {target_language} reader to fully grasp the intended message.
- 3: Good translation; the text is mostly fluent and conveys the original meaning well, but may have minor awkwardness or slight inaccuracies that could confuse a {target_language} reader.
- 4: Very good translation; the text is smooth and natural, effectively conveying the intended meaning, but may still have minor issues that could slightly affect understanding for a {target_language} reader.
- 5: Excellent translation; the text is fluent and natural, conveying the original meaning clearly and effectively, with no significant issues that would hinder understanding for a {target_language} reader.

Please provide the reason first, followed by a score. Format your evaluation in the XML structure below:
<evaluation>
<reason>
reason for the score
</reason>
<score>
float (0.00 to 5.00, two decimal places)
</score>
</evaluation>

[User]
<text>
{source}
</text>
<translation>
{translation}
</translation>
\end{tcblisting}
\end{figure*}

\section{Evaluation Prompts}
\label{app:eval_prompts}

The following prompts are used for the evaluation metrics reported in our experiments (\S\ref{sec:experiments}):
\begin{itemize}
    \item CRF (Appendix~\ref{app:crf})
    \item CEA5: uses the same prompt as the Evaluator (Appendix~\ref{app:evaluator})
    \item CEA100 (Appendix~\ref{app:cea100})
\end{itemize}

\begin{figure*}[t]
\subsection{CEA100 (Claude Evaluator Agent, 100-point scale)}
\label{app:cea100}
The CEA100 metric uses the same prompt structure as CEA5, with the scoring scale adjusted to 0--100:
\begin{tcblisting}{colback=gray!8, colframe=gray!40, listing only, listing options={basicstyle=\footnotesize\ttfamily, breaklines=true, breakatwhitespace=false, columns=flexible}}
[System]
Please evaluate the following {target_language} translation of an English text. Rate the translation on a scale of 0 to 100, where:
- 10 points: Poor translation; the text is somewhat understandable but contains significant errors and awkward phrasing that greatly hinder comprehension for a {target_language} reader.
- 30 points: Fair translation; the text conveys the basic meaning but lacks fluency and contains several awkward phrases or inaccuracies, making it challenging for a {target_language} reader to fully grasp the intended message.
- 50 points: Good translation; the text is mostly fluent and conveys the original meaning well, but may have minor awkwardness or slight inaccuracies that could confuse a {target_language} reader.
- 70 points: Very good translation; the text is smooth and natural, effectively conveying the intended meaning, but may still have minor issues that could slightly affect understanding for a {target_language} reader.
- 90 points: Excellent translation; the text is fluent and natural, conveying the original meaning clearly and effectively, with no significant issues that would hinder understanding for a {target_language} reader.

Please provide the reason first, followed by a score. Format your evaluation in the XML structure below:
<evaluation>
<reason>
reason for the score
</reason>
<score>
int
</score>
</evaluation>

[User]
<text>
{source}
</text>
<translation>
{translation}
</translation>
\end{tcblisting}
\end{figure*}

\begin{figure*}[t]
\subsection{CRF (Claude Reference-Free)}
\label{app:crf}
\begin{tcblisting}{colback=gray!8, colframe=gray!40, listing only, listing options={basicstyle=\footnotesize\ttfamily, breaklines=true, breakatwhitespace=false, columns=flexible}}
[System]
Score the following translation from English to {target_language} on a continuous scale from 0 to 100, where score of zero means "no meaning preserved" and score of one hundred means "perfect preservation of meaning, with faithfulness, expressiveness, and elegance".

Format your score in the XML structure below:
<score>
int
</score>

[User]
English source: {source}
{target_language} translation: {translation}
\end{tcblisting}
\end{figure*}

\end{document}